\documentclass[acmsmall,screen,journal]{acmart}

\usepackage{multirow}
\usepackage{algorithm}
\usepackage{algorithmic}
\usepackage{amsmath}
\usepackage[outdir=./]{epstopdf}

\usepackage{subfigure}
\usepackage{array}
\usepackage{booktabs}
\usepackage{colortbl}

\AtBeginDocument{%
  }

\setcopyright{acmcopyright}
\copyrightyear{2023}
\acmYear{2023}

\acmJournal{TIST}

\renewcommand\footnotetextcopyrightpermission[1]{}

\begin{document}

\title{Unifying Gradients to Improve Real-World Robustness for Deep Networks}

\author{Yingwen Wu}
\affiliation{%
  \institution{Institute of Image Processing and Pattern Recognition, Shanghai Jiao Tong University}
  \streetaddress{800 Dongchuan Road}
  \city{Shanghai}
  \country{China}
  }
\email{yingwen_wu@sjtu.edu.cn}

\author{Sizhe Chen}
\affiliation{%
  \institution{Institute of Image Processing and Pattern Recognition, Shanghai Jiao Tong University}
  \streetaddress{800 Dongchuan Road}
  \city{Shanghai}
  \country{China}
  }
\email{sizhe.chen@sjtu.edu.cn}

\author{Kun Fang}
\affiliation{%
  \institution{Institute of Image Processing and Pattern Recognition, Shanghai Jiao Tong University}
  \streetaddress{800 Dongchuan Road}
  \city{Shanghai}
  \country{China}
  }
\email{fanghenshao@sjtu.edu.cn}

\author{Xiaolin Huang}
\affiliation{%
  \institution{Institute of Image Processing and Pattern Recognition and the MOE Key Laboratory of System Control and Information Processing, Shanghai Jiao Tong University}
  \streetaddress{800 Dongchuan Road}
  \city{Shanghai}
  \country{China}
  }
\email{xiaolinhuang@sjtu.edu.cn}

\renewcommand{\shortauthors}{Yingwen Wu, Sizhe Chen, Kun Fang, Xiaolin Huang}


\begin{abstract}
  The wide application of deep neural networks (DNNs) demands an increasing amount of attention to their real-world robustness, \emph{i.e.}, whether a DNN resists black-box adversarial attacks, among which score-based query attacks (SQAs) are most threatening since they can effectively hurt a victim network with the only access to model outputs. Defending against SQAs requires a slight but artful variation of outputs due to the service purpose for users, who share the same output information with SQAs. In this paper, we propose a real-world defense by Unifying Gradients (UniG) of different data so that SQAs could only probe a much weaker attack direction that is similar for different samples. Since such universal attack perturbations have been validated as less aggressive than the input-specific perturbations, UniG protects real-world DNNs by indicating attackers a twisted and less informative attack direction. We implement UniG efficiently by a Hadamard product module which is plug-and-play. According to extensive experiments on 5 SQAs, 2 adaptive attacks and 7 defense baselines, UniG significantly improves real-world robustness without hurting clean accuracy on CIFAR10 and ImageNet. For instance, UniG maintains a model of $77.80\%$ accuracy under 2500-query Square attack while the state-of-the-art adversarially-trained model only has $67.34\%$ on CIFAR10. Simultaneously, UniG outperforms all compared baselines in terms of clean accuracy and achieves the smallest modification of the model output. The code is released at https://github.com/snowien/UniG-pytorch.
\end{abstract}


\maketitle

\section{Introduction}
Deep neural networks (DNNs) have been revealed to be vulnerable to adversarial examples (AEs), which can mislead models into incorrect predictions by imperceptible perturbations on inputs \cite{szegedy2013intriguing, goodfellow2014explaining,zhang2020adversarial}. Such sensitivity poses a threat to real-world applications of DNNs, since an attacker only needs the same information as the user, namely the model output, to generate valid AEs. Considering that the inner information of the network, such as parameters or training datasets, is hidden from attackers in practical scenarios, the query-based attack \cite{square,NES,signhunter,SimBA,Bandits, chen2021querynet,chen2020hopskipjumpattack, first-decision-based-attack}, which only 
requires the model output to hurt a victim model, deserves more attention in the field of real-world robustness compared to white-box attacks \cite{goodfellow2014explaining, carlini2017towards} and transferable attacks \cite{huang2019black, chen2020universal, chen2022relevance}. Moreover, since plentiful applications output both prediction results and probabilities \cite{bojarski2016end,bojarski2016end,caruana2015intelligible} to users for better judgement, score-based query attacks (SQAs) are more threatening because of their effectiveness and feasibility, compared to decision-based query attacks (DQAs), which needs unreasonable number of queries to attack.

Since SQAs are based on output probabilities to generate AEs, current defenses are all designed to alter the probabilities, either directly or indirectly, to resist attackers. For instance, adversarial training (AT), the most popular defense, uses on-the-fly AEs as training data to obtain a robust model whose output probabilities are always under-confident \cite{madry2017towards, tramer2017ensemble, gowal2021improving, li2022subspace}. Randomness injection (RI) is another effective defense against SQAs, which inject randomness into inputs \cite{cohen2019certified,xie2017mitigating,RND}, parameters \cite{liu2018adv,PNI,dio}, or features \cite{RSE} to ultimately change probabilities with random noises to confound attackers. In contrast to RI, denoising methods \cite{nie2022diffusion,alfarra2022combating,pang2019mixup} pre-process inputs to mitigate adversarial noises and reconstruct natural images, such that the output probability of adversarial queries can be modified to be consistent with the score of the corresponding clean query. Additionally, dynamic defenses \cite{RMC,DENT}, which optimize model parameters at inference time to adapt attacks, also try to keep the same output probabilities with clean data when adversarial queries come. Although above defenses could mitigate SQAs, however, they hurt clean accuracy, which is a common phenomenon called accuracy-robustness trade-off \cite{engstrom2019robustness,pinot2019theoretical,raghunathan2019adversarial,zhang2019theoretically,xie2020adversarial}. An intuitive explanation to this trade-off is that high clean accuracy requires the use of detailed features, while they are the inducement of vulnerability \cite{engstrom2019robustness}. From distribution perspective, natural images and adversarial examples belong to different distributions and thus cannot be fitted well simultaneously, which is simply proved by AT. In addition to the degradation of clean accuracy, we discover that these defenses inevitably affects the output probabilities of clean images, which seriously influences downstream tasks to make reasonable decisions, \emph{e.g.}, the detection network needs to report accurate confidence to the center controller to avoid erroneous decisions on the object with low confidence. Therefore, in this paper, we aim at keeping accuracy and probabilities of clean data and meanwhile changing output probabilities of adversarial queries to simultaneously serve users and resist attackers. 

Achieving the above goal is difficult because that both users and attackers share the same information, \emph{i.e.}, the model output probability, while we need to keep the probability of clean data and change the probability of adversarial queries in the condition of unknown input types. In spite of this, our chance lies in the fact that users only ask for outputs, while SQAs concentrate on the change in output indicated by different queries, which implies gradient information used to attack. Therefore, we propose a defense that explicitly changes the gradient information contained in the output of consecutive queries. Through slight but designed modifications on outputs, which guarantee the service to the user, the gradient information contained in the output is perturbed to be an elaborate direction which is less aggressive. We choose the direction of the universal attack perturbation (UAP, \cite{ijcai2021p635}) here, which is consistent for different inputs and hereby less threatening than normal adversarial noises that are image-specific. Previous studies have proved the gentleness of UAP empirically \cite{zhang2021data,zhang2020cd,ijcai2021p635,benz2020double}. As a result, even through the attacker mines the attack direction from these outputs, their attack trajectory will be tricked away from the vulnerable adversarial direction and induced into the weaker path we have designated.

\begin{figure}[h!]
    \centering
    \subfigure[Method flowchart]{
    \includegraphics[width=0.45\textwidth,height=0.25\textwidth]{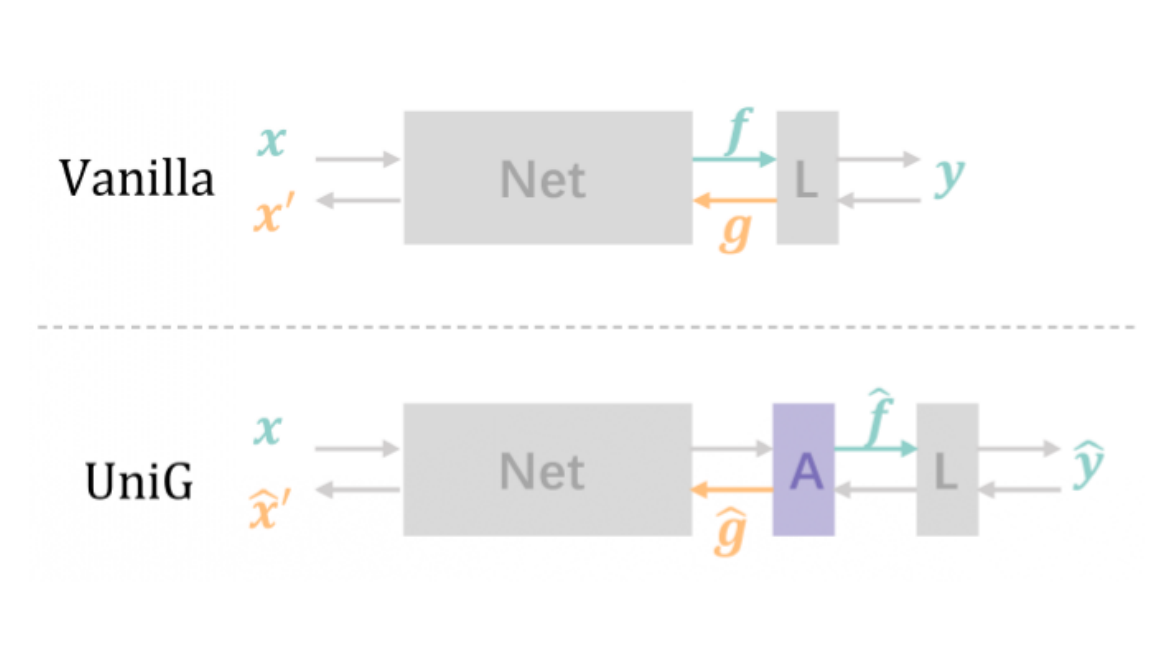}
    \label{flowchart}
    }
    \subfigure[Output changes and robustness]{
    \includegraphics[width=0.45\textwidth,height=0.25\textwidth]{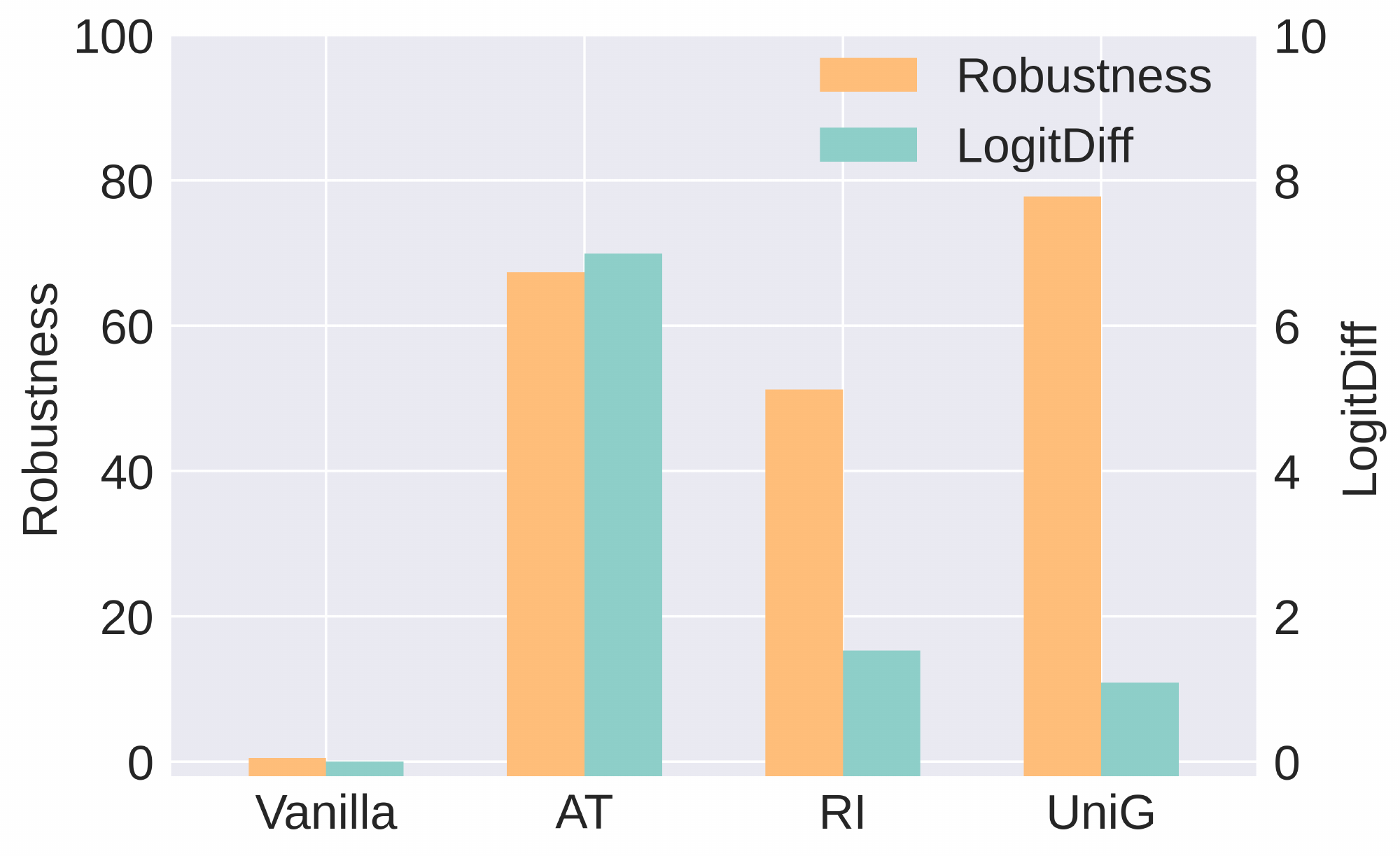}
    \label{trade-off}
    }
    \caption{(a) The flowchart of our UniG method. The plug-in Hadamard product module $A$ slightly modifies the forward feature ($\hat{f}\approx f$) but totally distorts the backward gradient ($\hat{g}\neq g$), where $\hat{g}$ is a less threatening attack direction. (b) The robust accuracy (under a SOTA SQA named Square \cite{square}) and the logit (output) difference on clean data. Our UniG has both the best robustness and smallest output modifications compared with SOTA AT \cite{gowal2021improving} and RI \cite{RND} defenses.}
\end{figure}

To confuse SQAs towards the UAP direction, the modification on outputs needs to be designed carefully. A natural idea is that we use a gradient unification loss, which constrains the gradient of different inputs to be the same, to optimize the output changes. The proposed method is then called unifying gradients (UniG) approach. Considering computational overhead, we choose to unify the gradient of features instead of inputs to efficiently calculate the second derivative in practice, as shown in Figure \ref{flowchart}. We insert a Hadamard product module $A$ into a pre-trained DNN, where we want $\hat{f}\approx f$ for user friendliness and $\hat{g}\neq g$ but rather a UAP direction for adversarial robustness. The specific calculation of $\hat{f},\hat{g}$ is that $\hat{f}=A\circ f$, $\hat{g}_i=g_i\circ A_i$ ,where $A\in R^{b\times d}$, $A_i\in R^{d}$ ($b$ for batch size and $d$ for feature dimension) is the module parameter and $g_i\in R^{d}$ is the feature gradient of the $i$-th input. According to the design of $\hat{f}$ and $\hat{g}$, we constrain each element of $A$ to be close to one to ensure slight forward modifications, and optimize $A$ by minimizing the gradient variance between different images to transform $\hat{g}$ to be a UAP direction. As a result, our method conceals the vulnerable attack direction $g$ and instead displays a weaker one $\hat{g}$ through the output probabilities $\hat{y}$ affected by the optimized module $A$. The advantages of our approach are listed below.

\begin{itemize}
    \item The module $A$ is plug-and-play for any pretrained network with negligible additional computation. Each time a test batch arrives, according to our experiments, the module parameters can be well re-optimized from random initialization within one epoch using the current test data (Noting that the test data is unlabeled, we compute the cross-entropy loss by using the predicted class as the label). UniG is remarkably lightweight compared to other dynamic defenses \cite{RMC,DENT} that optimize the entire model in inference time.
    \item In contrast to other defenses, our UniG performs best in improving real-world robustness with the least modification on output probabilities, see Figure \ref{trade-off}. Our explicit optimization objective of changing the gradient information contained in the output fundamentally suppresses the attack performance of SQAs, while other defenses such as RI randomly change the forward output to indirectly perturb the attacker's estimation on the gradient.
\end{itemize}

We compare our UniG with seven defenses \cite{gowal2021improving,salman2020adversarially,RND,PNI,DENT,pang2019mixup,alfarra2022combating,nie2022diffusion} on CIFAR10 \cite{cifar10} and ImageNet \cite{imagenet} under five popular SQAs \cite{square,SimBA,signhunter,NES,Bandits}. The result shows that our UniG can keep clean accuracy ($94.26\%$ on CIFAR10 \cite{cifar10} and $78.47\%$ on ImageNet \cite{imagenet}) while obtaining the best robustness in most cases (the remaining accuracy under 2500-query Square attack is $77.80\%$ on CIFAR10, while the state-of-the-art (SOTA) AT model only achieves $67.34\%$). Additionally, under adaptive attacks in black-box scenarios such as steal-based \cite{kariyappa2021maze,orekondy2019knockoff} and hyperparameter-tuning SQAs \cite{RND}, our UniG also protects DNNs best compared to other defenses \cite{RND,DENT}. Note that UniG as a black-box defense, while modifying gradients, does not belong to obfuscated gradients defenses \cite{athalye2018obfuscated} which resist white-box attacks effectively but cannot mitigate black-box ones.

\section{Related Work}
\subsection{Score-based Query Attacks}
Current SQAs can be categorized into two types: gradient estimation \cite{signhunter,NES,Bandits,zoo} and random direction search \cite{square,SimBA}. For the first type, attackers calculate directional derivation using the output of continuous queries to estimate the input gradient \cite{zoo}. Based on this idea, Bandits \cite{Bandits} further reduces query times through utilizing data-dependent gradient prior information. Natural Evolutionary Strategies (NES, \cite{NES}) considers the condition of limited information, eg. only top-k probabilities are accessible, and develops the corresponding SQA. SignHunter \cite{signhunter} constructively proposes to ignore gradient magnitudes and only focus on gradient signs to improve attack efficiency. For the second type, attackers randomly choose an attack direction and adjust or directly abandon it according to the feedback from models. Square \cite{square} is the most popular approach in this category, which adds localized square-shape random noises into inputs. In addition to Sqaure, SimBA \cite{SimBA} is an early and simple random-search attack, which selects perturbations from orthonormal basis.

\subsection{Adversarial Defense}
Before, adversarial defenses are mainly designed to resist white-box attacks \cite{szegedy2013intriguing,carlini2017towards,madry2017towards}, among which adversarial training (AT \cite{tramer2017ensemble,gowal2021improving}) is the most popular and comprehensive. 
Recently, defenses for SQAs are proposed in quantity, which can be categorized into three types: adversarial detection, denoising and randomness injection. The first one utilizes the similarity between malicious query data to detect adversarial examples \cite{pang2020advmind,li2020blacklight,chen2020stateful}. However, it takes large storage resources and meets problem when facing long-interval queries. The basic idea of denoising methods is to pre-process inputs to eliminate adversarial perturbations \cite{nie2022diffusion,alfarra2022combating,pang2019mixup}. Consequently, outputs for continuous queries remain unaltered, so that SQA attackers cannot obtain valid information of gradients. For instance, \cite{nie2022diffusion} proposes to use a pre-trained diffusion model as the input denoising network. \cite{alfarra2022combating} adds inverse adversarial perturbations to inputs, and \cite{pang2019mixup} mixups the input with other random clean samples, all for the purpose of shrinking the adversarial perturbation. In contrast, randomness injection (RI) approaches aim to bewilder attackers by random noises on the outputs. To achieve this, they inject random noises into different parts of models such as inputs \cite{byun2022effectiveness,RND,xie2017mitigating}, features \cite{RSE} and parameters \cite{PNI}. In addition, \cite{cohen2019certified,lecuyer2019certified,salman2019provably} propose to combine randomness into training process for certified robustness. Although randomness protects models from attacking, it inevitably reduces clean accuracy. Apart from the above defenses, recently proposed transductive methods \cite{RMC,DENT} are highly related to our method, which dynamically optimize network parameters at test time to adapt attacks. Nevertheless, the test speed of dynamic defenses is slow because of their optimization process at each inference time. Different from them, our method only needs to adjust the parameter of our designed module instead of the whole network, thus is more practical. Although our approach, like the others, defends against SQAs by changing the output probabilities, the output modification in our method is obtained via optimizing a designed goal of keeping forward information and concealing backward knowledge, instead of random noises. This is why we obtain better accuracy and defense performance compared to RI methods which do not have a clear optimization goal.


\section{Method}
\subsection{Preliminaries and Motivation}
Let us denote the victim model as $M:X\rightarrow Y$, and the benign data as $(x,y)\in (X,Y)$. An attack aims to craft an adversarial example $x'$ which locates at the neighborhood of $x$ but misguides the model to an incorrect prediction class. The attack optimization problem can be summarized as
\begin{equation}\label{attack}
    \begin{aligned}
       &\min_{x'\in N_r(x)} l(x') = \min_{x'\in N_r(x)}  (M_{y}(x')-\max_{j\ne y} M_{j}(x')),\\
    \end{aligned}
\end{equation}
where $N_r(x)=\{x'|\Vert x-x'\Vert_p \leq r\}$ indicates the $l_p$ ball around $x$ with a radius $r$, and $M_{y}$ denotes the predicted logits (or probabilities) of the $y$-th class. An effective attack algorithm can make $l(x')\le 0$ so that the predicted class is no longer the true label $y$. A common method to solve (\ref{attack}) is to iteratively optimize the objective function by gradients of $l(x')$ \emph{w.r.t.} $x'$, namely projected gradient descent (PGD, \cite{madry2017towards}) algorithm.

However, for SQAs, the gradient cannot be obtained directly since only the output probability of the victim model is accessible to attackers. Therefore, SQAs use direction derivation or random search method to estimate the gradient. The key of these methods is to utilize the forward output of queries to infer backward gradient information. Accordingly, we propose a defense idea that modifies the forward outputs slightly to prevent attackers from estimating backward gradients. 

\subsection{UniG: Unifying Data's Gradients to Defend against SQAs}
We intend to change the gradient information contained in the output of queries to fool the attacker into a distorted attack trajectory, and simultaneously keep the output of clean data as far as possible. The overall goal of our method can be summarized as follows:
\begin{eqnarray}\label{2}
\begin{cases}
	\hat{M}(x) \approx M(x)\\
	\hat{G}(x) \neq G(x)
\end{cases}
\end{eqnarray}
where $M(x),\hat{M}(x)$ denote the output logits (probabilities) of the vanilla and defense model respectively, and $G(x),\hat{G}(x)$ are the backward gradient of inputs of the vanilla and defense model. We use the gradient constraint to guide our slight modification on outputs. The gradient information contained in our slightly modified output $\hat{M}(x)$ is the artful direction $\hat{G}(x)$ instead of the true vulnerable trajectory $G(x)$. The SQA attacker hereby can only dig an attack direction $\hat{G}(x)$ from the output of queries to generate AEs, which is designed to be weaker than $G(x)$. Here, we choose $\hat{G}(x)$ to be the direction of the universal attack perturbation \cite{ijcai2021p635} which is consistent between different samples and proved to be weaker \cite{zhang2021data,zhang2020cd,ijcai2021p635,benz2020double} than normal attack directions which are image-specific \cite{madry2017towards}. According to the above discussion, we propose an optimization problem to generate our modification on outputs to defend against SQAs.
\begin{equation}\label{3}
    \begin{aligned}
       &\min_{\hat{M}} &&\sum_{i=1}^{n} (\hat{G}(x_i)-\hat{G}(x_{i+1}))^2\\
       & \mathrm{s.t.} && \Vert \hat{M}(x)-M(x) \Vert \leq \delta,
    \end{aligned}
\end{equation}
The constraint obviously corresponds to the slight modification target, where $\delta$ controls the degree of output offset. The objective function corresponds to the goal of distorting the gradient information into a universal one which is less threatening, where $\hat{G}(x_i)$ represents the gradient of the $i$-th input. 

To solve problem (\ref{3}), a direct way is to finetune the victim model using the above loss function. However, this solution is not only resource-consuming, but also difficult because of millions of network parameters and training samples. Therefore, we propose an alternative solution. We replace $\hat{M}$ with a simple module $A$, which is inserted into the penultimate layer of the victim model as shown in Figure \ref{flowchart}. The parameter of the victim model remains unchanged and only the module $A$ needs optimization. Moreover, we choose to unify the gradient of features instead of inputs to simplify the optimization process. The overall problem becomes the following formulation,
\begin{equation}\label{4}
    \begin{aligned}
       &\min_{A} &&l(A,x) = \sum_{i=1}^{b-1} (\hat{g}(x_i) - \hat{g}(x_{i+1}))^2\\
       & \mathrm{s.t.} &&\Vert \hat{f}-f\Vert \leq \delta.
    \end{aligned}
\end{equation}
where $\hat{f}$, $\hat{g}$, and $b$ respectively denote feature, feature gradients, and batch size as shown in Figure \ref{flowchart}. The operation of $A$ is designed to do Hadamard product with input features, that is, $\hat{f}=A \circ f$, which is simple but effective. To keep forward outputs, every element of $A$ is expected to be close to one, while the objective loss in problem (\ref{4}) gives an instructional direction to optimize $A$ around all-one matrix. As a result, the slight modification on features plays an important role in distorting backward information to mislead attackers.

We solve problem (\ref{4}) using gradient descent (GD) algorithm, and the process of optimizing $A$ is integrated into the forward calculation of our defense model, see Alg. \ref{alg:algorithm1}. 
At each inference time, the module $A$ is re-optimized from random Gaussian initialization using current test data with the objective function in Eq. (\ref{4}). Noticing that test data have no labels, we utilize the prediction label as the true label in cross-entropy loss to calculate $\hat{g}(x)$. 
It is worth noticing that the objective loss is easy to optimize because only the linear layer and our designed module are involved in the calculation. Our experiment results show that with only one iteration step, the gradient variance can be efficiently minimized to significantly improve the robustness under SQAs. 
\begin{algorithm}[htp]
  \caption{Forward Calculation of UniG Model (The symbols correspond to those in Figure \ref{flowchart})}
  \label{alg:algorithm1}
  \begin{algorithmic}[1]
    \REQUIRE{batch data: $x$; optimization iterations: $p$; learning rate: $\alpha$; constraint parameter: $\delta$}
    \ENSURE{model prediction: $\hat{y}$}
	\STATE Initialize parameters $A$ with $A\sim \mathcal N(1,0.5)$
	\FOR{$i=1$ to $p$}
        \STATE Compute $f=Net(x)$, $y=L(f)$, $\hat{y}=L(A(f))$, $c=OneHot(y)$, $CE\_loss=-\sum c_i\cdot log(\hat{y}_i)$
    	\STATE Calculate and min-max normalize the gradient $\hat{g}$
    	\STATE Compute the objective loss $l(A,x)$ in problem (\ref{4})
        \STATE Update $A$ with $A\leftarrow A - \alpha\cdot\frac{\partial l(A,x)}{\partial A}$
        \STATE Clip $A$ with $\Vert A_{ij}-1\Vert_{\infty} \leq \delta$
    \ENDFOR
    \STATE Compute final output $\hat{y}=L(A(f))$ with optimized $A$
    \RETURN $\hat{y}$	
    \end{algorithmic}
\end{algorithm}

\subsection{Discussion}

\begin{figure}[h!]
    \centering
    \subfigure[Objective function value \emph{w.r.t} iteration step]{
    \includegraphics[scale=0.3]{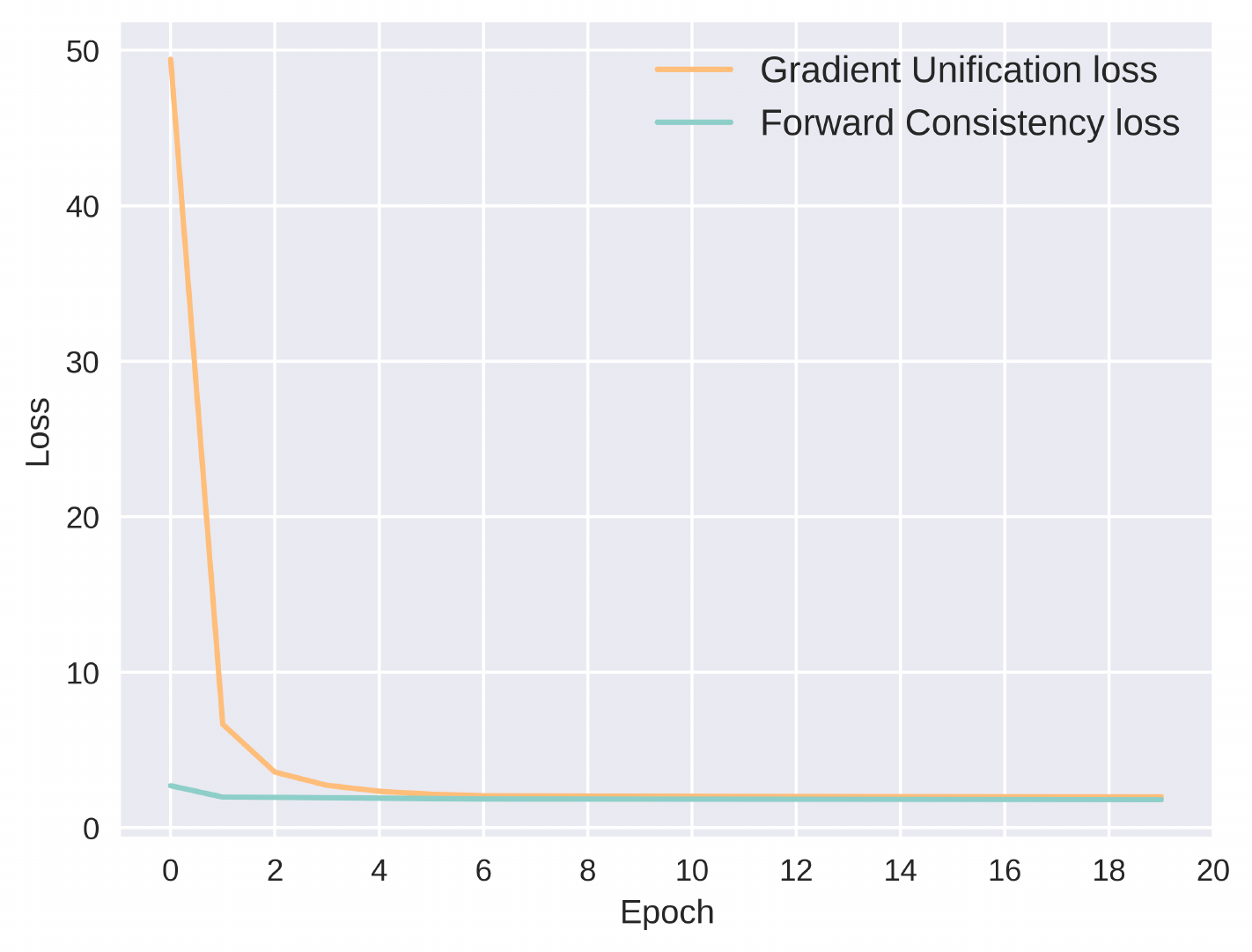}
    \label{training-loss}
    }
    \qquad
    \quad
    \subfigure[Margin loss \emph{w.r.t} Query times]{
    \includegraphics[scale=0.3]{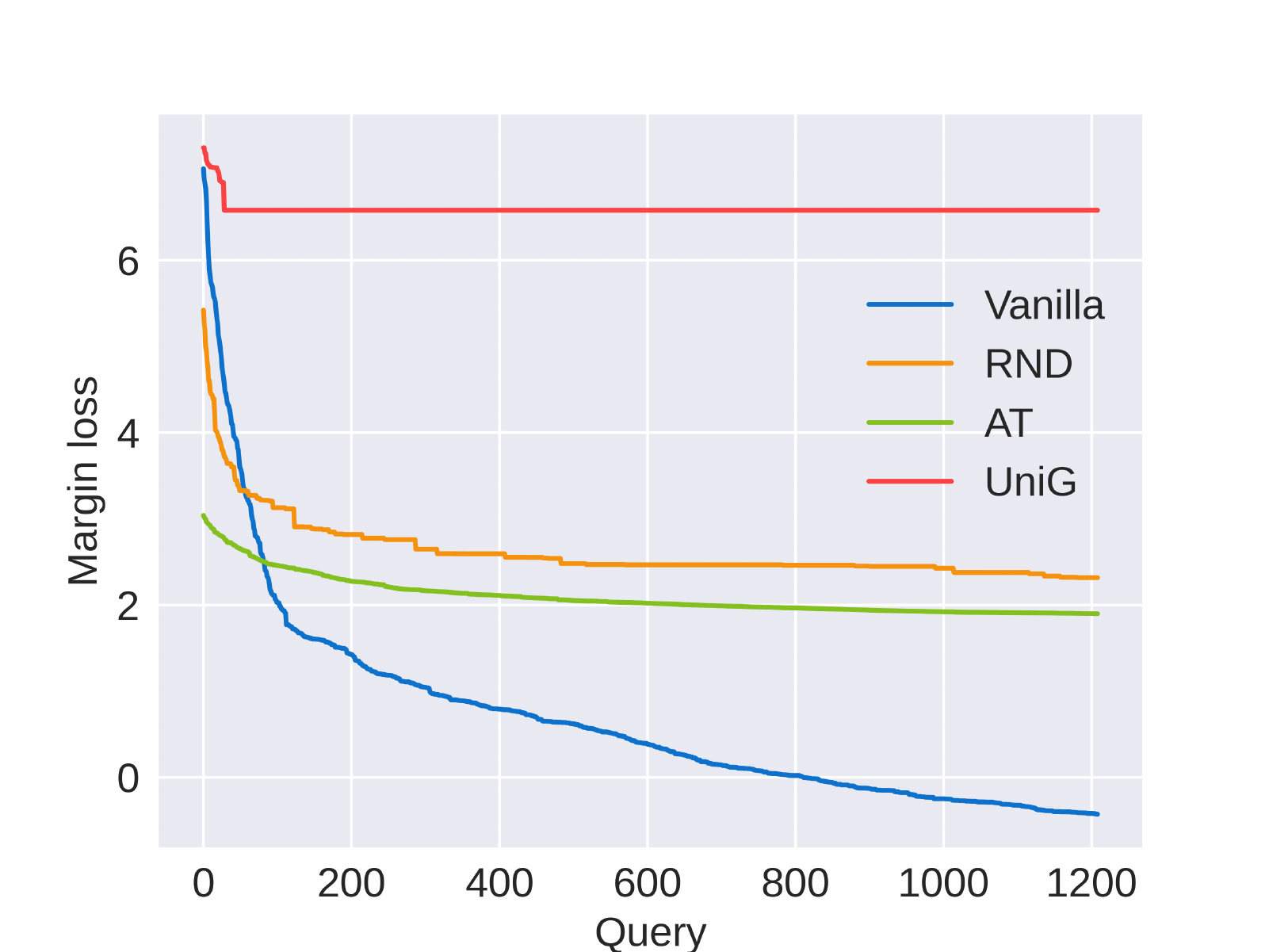}
    \label{margin-loss-query}
    }
    \caption{(a) Objective function value (UniG Loss in Eq. (\ref{4})) \emph{w.r.t} iteration step of our defense model. Since each forward calculation the loss is re-optimized with the current test data, we randomly choose one forward process to present the loss change. (b) Margin loss \emph{w.r.t} Query times of four compared defenses under Square attack. A higher loss means a more robust network.}
\end{figure}

\begin{figure}[h!]
    \centering
    \subfigure[Adversarial examples and noises of UniG model]{
    \includegraphics[scale=0.24]{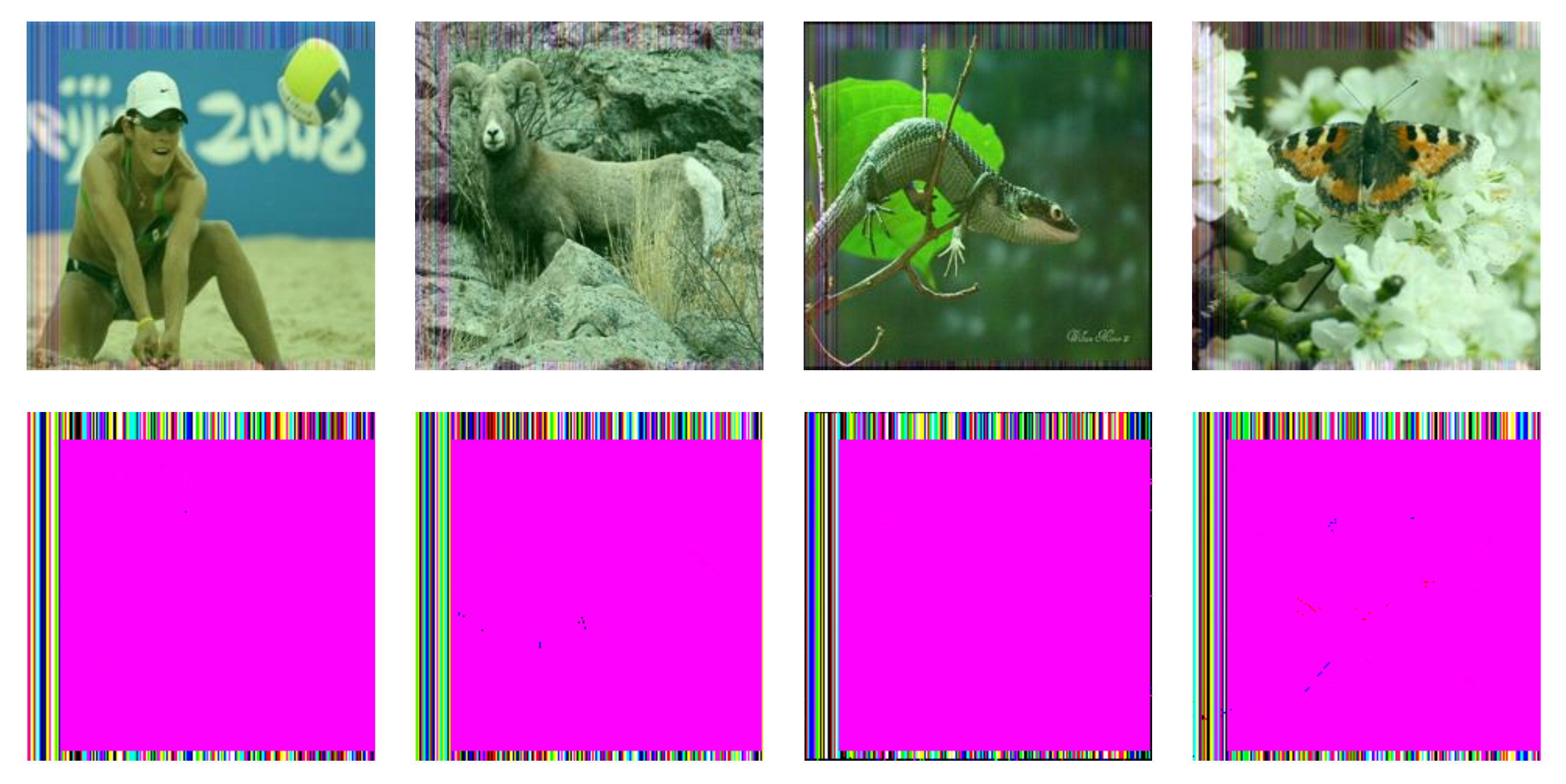}
    \label{uap}
    }
    \subfigure[The partial value of the matrix $A$ for different inputs]{
    \includegraphics[scale=0.22]{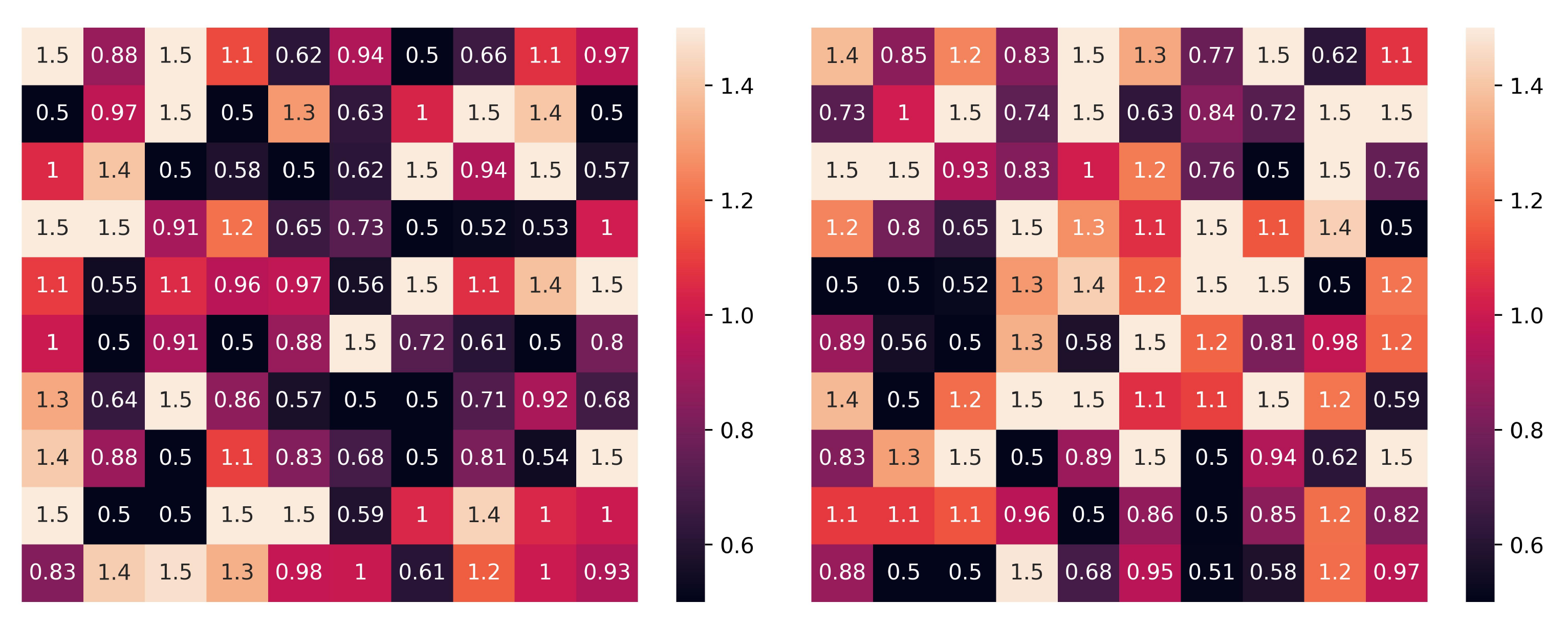}
    \label{a_matrix}
    }
    \caption{(a) Adversarial perturbations and the corresponding AEs of our UniG model for different images in one test batch, where the perturbations are the same for diverse images. (b) The partial value of the matrix $A$ for different inputs (queries), indicating that different data lead to divergent values of the matrix $A$.}
\end{figure}

To further elucidate our method, we display the training process and the final value of $A$ as showing in Figure \ref{training-loss},\ref{a_matrix}. From Figure \ref{training-loss}, we observe that the gradient unification loss is well-optimized with one iteration step (The forward consistency loss is calculated by $\Vert \hat{f}-f\Vert$ to show the modification on forward features is tiny.). Figure \ref{a_matrix} shows partial final values of the matrix $A$ (the parameter of the designed module), which demonstrates that although randomly initialized, the value of $A$ is highly dependent on the current batch data after optimization, thus is different for each query. The difference is because that gradients are naturally diverse for different data, so unifying them requires divergent values of the matrix $A$.

Since our method claims that we fool attackers into a universal attack direction, we visualize the AEs and adversarial perturbations of our UniG model to verify it. The result in Figure \ref{uap} directly illustrates our statement, where the adversarial noise of our UniG model keeps consistent for different images, while that of the vanilla network obviously is different for diverse images. Because of the distortion of image-dependent attack directions, in our defense, the marginal loss in Eq. (\ref{attack}) hardly decreases as the number of queries increases, see Figure \ref{margin-loss-query}.

Considering a possible situation where the model receives a single image to test, our approach, like other dynamic defenses based on batch optimization, needs a solution to deal with this condition. One possible way is that we could cascade several training data with the test sample to perform such optimization for the single test sample situation. We specifically discuss it with experiment results in Section \ref{Other Algorithmic Discussions}.

\section{Experiment\label{experiment}}
\subsection{Setup}
\textbf{Datasets.}
We experiment on two widely-used classification datasets: CIFAR10 \cite{cifar10} and ImageNet \cite{imagenet}. For CIFAR10, the whole testset is used to evaluate method performances. For ImageNet, we randomly choose 1K images (one image for one class) from the validation set as test data. 

\textbf{Models.}
PreResNet18 \cite{preresnet} and WideResNet-50-2 \cite{wideresnet} are used as typical model architectures. 
On CIFAR10, PreResNet18 with vanilla training achieves $94.26\%$ accuracy. On ImageNet, we directely use the pre-trained model from torchvision package of PyTorch \cite{paszke2019pytorch}, which has $78.47\%$ accuracy. 

\textbf{Score-based Query Attacks.}
We evaluate defense methods under five 
popular SQAs, including Square \cite{square}, SimBA \cite{SimBA}, Sign \cite{signhunter}, NES \cite{NES}, and Bandits \cite{Bandits}. The first two are based on random search and the rest are based on gradient estimation. Most of the results in this paper are obtained under the untargeted  $\ell_\infty$ norm attack, but we also present the results of the targeted and $\ell_2$ norm attack in Section \nameref{more_attacks} for completeness. The $\ell_\infty$ attack bound is set as $8/255$ for CIFAR10 and $4/255$ for ImageNet, and the $l_2$ attack bound is set as $1$ for CIFAR10 and $5$ for ImageNet. The query budget of SQAs is set to be 100 and 2500 for evaluating defense methods under different attack intensities. 

\textbf{UniG setting.}
The UniG module is plugged into vanilla training models. For its parameters, we set $\delta$=0.5, $p$=1, $\alpha$=10 on CIFAR10 and $\delta$=0.1, $p$=1, $\alpha$=1 on ImageNet. 

\textbf{Compared defenses.}
The proposed defense method will be compared with adversarial training (AT, \cite{gowal2021improving,salman2020adversarially}), random noise defense (RND, \cite{RND}) and its enhancement (RND-GF), parameter noise injection (PNI, \cite{PNI}), dynamic inference (DENT, \cite{DENT}), mixup inference (Mixup, \cite{pang2019mixup}), Anti-adversary combination (Anti-adv, \cite{alfarra2022combating}) and adversarial purification based on Diffusion models (DDPM, \cite{nie2022diffusion}) approaches. The first one is the most popular defense, the second to fourth ones belong to RI, the fifth one is a dynamic defense, and the rest are denoising methods. The AT models are obtained from Robustbench\footnote{https://robustbench.github.io/} \cite{croce2021robustbench} and we choose the most robust one for comparison. Random noise defense, which adds random noise into inputs or outputs, is conducted with noise variance $0.02$, as recommended by \cite{RND}. RND-GF is fulfilled by fine-tuning the baseline model with $100$ epochs, and then we test it using variance $0.05$. The model of PNI is obtained from the corresponding github code\footnote{https://github.com/elliothe/CVPR$\_$2019$\_$PNI} and we choose the best for comparison. DENT is a transductive method and we insert it into the vanilla training model, as the same as UniG. We try different mixup ratios and choose the best one $0.9$ to combine Mixup\footnote{https://github.com/P2333/Mixup-Inference} method with 
our baseline and AT model. For Anti-adv\footnote{https://github.com/MotasemAlfarra/Combating-Adversaries-with-Anti-Adversaries}, we use the recommended iteration number $K=2$ and the anti-adversary step is set to $4/255$ for the best performance since the common test attack step is $8/255$. We directly utilize the pre-trained diffusion model in DDPM code\footnote{https://github.com/NVlabs/DiffPure} to denoise inputs for each query.

\textbf{Metric.}
For a good defense, we need to consider three folds: the clean accuracy, the logit difference, and the robust accuracy, \emph{i.e.}, the remaining accuracy under attacks. 
The logit difference reflects the output difference between vanilla model and the current defense model, whose value is highly proportional to the probability difference and more pronounced to observe. In this paper, we use $l_2$ norm to measure it. Other norms can be used as well and the conclusion is similar. The smaller it is, the more friendly the model is to users. 

\subsection{Defense Performance}
Table 1 comprehensively reports the defense performance of the proposed UniG together with AT, RND, RND-GF, PNI, DENT, Mixup, Anti-adv and DDPM. Due to the huge computing overhead of DDPM, we just test its performance under the Square attack which is classical and popular. AT could improve the robust accuracy under different attacks but it degrades clean accuracy by about $7\%$ for CIFAR10 and $10\%$ for ImageNet. The other defenses undertake effort to avoid the downgrade on clean accuracy, but the robust accuracy is generally lower than AT, especially on CIFAR10. Although DENT performs well on ImageNet under various SQAs, it sacrifices the accuracy of output probabilities. In fact, we discover that the output probability of DENT model is almost composed of zeros or ones, making SQAs degrade into DQAs and hereby achieving good performance. Another defense named Mixup also outperforms our approach in some cases on ImageNet, however, it requires thirty additional forward calculations to perform its stochastic input mixing and output ensemble averaging, which adds too much computation burden in practice. By contrast, our UniG model achieves similar defense performance as AT, actually for most cases UniG is more robust than AT, and meanwhile greatly remains the clean accuracy, almost the same as the vanilla training. For logits difference, the advantage of UniG is more significant that the logits difference is generally reduced by an order of magnitude, which means UniG is much more user-friendly than the other defenses. 
\begin{table}
  \caption{The comparison of different defense methods under $l_\infty$ norm attacks (query = 100/2500) on CIFAR10 and ImageNet. The clean accuracy, logits difference and robust accuracy are reported. The higher the clean accuracy and robust accuracy, the smaller the logits difference, indicating the better performance of the defense method. The best results are in bold, and the 2nd ones are underline.}
  \label{comparison}
  \resizebox{\textwidth}{!}{
  \begin{tabular}{c|c|c|c|c|c|c|c|c}
    \toprule
    Datasets & Methods & Clean & Logit-diff & Square & SimBA & Sign & NES & Bandits\\
    \midrule
    \multirow{6}{*}{CIFAR10} & Vanilla & \underline{94.26} & - & 38.79/0.46 & 41.03/0.45 & 48.25/0.26 & 75.28/10.06 & 68.81/25.92 \\
     & AT & 87.35 & 7.00 & 79.15/67.34 & \underline{83.61}/71.36 & \textbf{78.28}/\underline{64.43} & 85.30/\underline{79.49} & \textbf{83.90}/74.33\\
     & RND & 91.14 & \underline{1.53} & 65.04/51.22 & 74.88/63.07 & 64.71/51.95 & 85.67/69.27 & 67.79/58.27 \\
     & RND-GF & 92.87 & 1.81 & 78.02/\underline{69.08} & 82.67/\underline{75.45} & 72.44/63.15 & 89.16/\textbf{81.73} & \underline{82.07}/\underline{74.89} \\
     & PNI & 85.93 & 8.16 & 64.66/51.54 & 64.13/57.41 & 65.31/50.48 & 80.77/69.30 & 66.93/56.81 \\
     & DENT & 94.25 & 7.35 & \underline{81.78}/57.71 & 75.23/54.00 & 64.09/46.18 & 88.60/67.80 & 74.55/69.75 \\
     & Mixup & \textbf{94.32} & 8.43 & 76.97/37.72 & 75.50/54.82 & 51.88/15.09 & \textbf{90.55}/75.46 & 78.91/73.96 \\
     & Anti-adv & 92.63 & 3.23 & 62.45/30.31 & 57.70/31.54 & 52.80/29.64 & 86.14/58.36 & 70.55/66.05 \\
     & DDPM & 88.88 & 2.30 & 52.50/42.80 & - & - & - & - \\
     & UniG & \underline{94.26} & \textbf{1.09} & \textbf{81.90/77.80} & \textbf{89.79/86.42} & \underline{72.58}/\textbf{68.81} & \underline{89.55}/67.88 & 80.31/\textbf{75.86} \\
    \midrule
    \multirow{5}{*}{ImageNet} & Vanilla & \underline{78.47} & - & 52.71/6.70 & 64.66/6.28 & 51.00/11.77 & 68.27/59.64 & 67.64/35.23 \\
     & AT & 68.41 & 49.23 & 62.25/51.92 & 67.50/51.92 & \underline{60.88}/\underline{56.10} & 65.67/65.67 & 66.49/61.64 \\
     & RND & 77.14 & \underline{13.73} & 61.13/48.44 & 71.22/65.25 & 59.40/54.77 & 74.83/\underline{72.51} & 70.02/67.42 \\
     & DENT & \textbf{78.60} & 98.73 & \underline{68.85}/\textbf{63.47} & \underline{77.69}/\underline{77.16} & \textbf{70.74}/\textbf{59.74} & \underline{77.03}/\textbf{74.67} & \textbf{76.62}/\textbf{74.25} \\
     & Mixup & 77.95 & 62.06 & \textbf{69.75}/\underline{54.57} & 77.65/- & 68.60/- & 76.39/- & \underline{75.44}/- \\
     & Anti-adv & 72.92 & 49.76 & 50.04/28.40 & 62.04/53.27 & 53.23/42.29 &  69.27/61.40 & 63.44/61.89 \\
     & UniG & \underline{78.47} & \textbf{2.75} & 66.14/52.88 & \textbf{78.22}/\textbf{77.44} & 57.28/45.51 & \textbf{77.68}/71.40 & 74.14/\underline{72.65} \\
  \bottomrule
\end{tabular}}
\end{table}

As a plug-and-play module, UniG can be readily applied in any network, for example in a model trained by adversarial examples.
Besides, since Mixup approach is based on AT models, we also conduct it to compare our performance here. The performance of UniG and Mixup in AT can be found in Table \ref{UniG-AT}. From AT, the classification accuracy of UniG-AT is well kept on clean examples and only drops $3.04\%$ on CIFAR10 under 2500-query Square attack, while the AT model drops about $20\%$. Similar performance could be found on both CIFAR10 and ImageNet under different attacks. Besides, UniG surpasses Mixup in terms of robustness both on CIFAR10 and ImageNet. The improvement of the Mixup method is unstable, seeing the drop in robust accuracy on ImageNet.
\begin{table}[htp]
  \caption{The performance of AT-based UniG model under SQAs (query = 100/2500) on CIFAR10 and ImageNet.}
  \label{UniG-AT}
    \resizebox{\textwidth}{!}{
  \begin{tabular}{c|c|c|c|c|c|c|c}
    \toprule
    Datasets & Methods & Clean & Square & SimBA & Sign & NES & Bandits\\
    \midrule
    \multirow{2}{*}{CIFAR10} & AT & 87.35 & 79.15/67.34 & 83.61/71.36 & 78.28/64.43 & 85.30/79.49 & 83.90/74.33\\
     & Mixup-AT & 86.53 & \textbf{84.91}/82.56 & 86.59/86.29 & 79.61/77.88 & 85.66/84.80 & \textbf{84.71}/\textbf{83.99}\\
     & UniG-AT & \textbf{87.36} & 84.61/\textbf{84.32} & \textbf{86.98/86.98} & \textbf{83.87/83.87} & \textbf{87.36/86.49} & 75.47/73.03 \\
    \midrule
    \multirow{2}{*}{ImageNet} & AT & \textbf{68.41} & 62.25/51.92 & 67.50/51.92 & 60.88/56.10 & 65.67/65.67 & 66.49/61.64 \\
     & Mixup-AT & 67.79 & 59.26/52.20 & 60.54/- & 64.40/- & 67.40/- & 67.34/- \\
     & UniG-AT & 68.36 & \textbf{65.49/63.71} & \textbf{68.29/68.09} & \textbf{64.94/64.94} & \textbf{68.36/67.68} & \textbf{67.60/66.92} \\
  \bottomrule
\end{tabular}}
\end{table}

\subsection{Performance under More Attacks\label{more_attacks}}
Before, we have shown the performance under current popular SQAs in an untargeted setting and a $\ell_\infty$-norm bound. In this section, we further evaluate the performance of our method under other settings, \emph{i.e.}, $\ell_{2}$ / $\ell_\infty$ norm and target (-T) / untarget (-UT) attacks as listed in the first row of Table \ref{generalization}. We adopt Square attack, the current SOTA SQA attack, for robustness evaluation. As Table \ref{generalization} verified, plugging UniG into vanilla and AT model can both achieve significant robustness improvement with a little clean accuracy drop. For instance, on CIFAR10, our UniG-Vanilla model improves robust accuracy by no less than $71.29\%$ under all attack settings with 2500 queries, and UniG-AT improves by $\geq 13.99\%$ compared with AT model, while our method keeps the same or even higher clean accuracy.
\begin{table}[htp]
  \caption{UniG under $\ell_{2}$/$l_\infty$ norm and target (-T) /untarget (-UT) Square attack (query = 100/2500)}
  \label{generalization}
  \begin{tabular}{c|c|c|c|c|c|c}
    \toprule
    Datasets & Methods & Clean & $\ell_{\infty}$-T & $\ell_{\infty}$-UT & $\ell_{2}$-T & $\ell_{2}$-UT\\
    \midrule
    \multirow{4}{*}{CIFAR10} & Vanilla & 94.26 & 45.24/0.35 & 38.79/0.46 & 56.56/6.60 & 62.21/5.66\\
     & UniG & \textbf{94.26} & \textbf{78.26/71.64} & \textbf{81.90/77.80} & \textbf{85.78/84.83} & \textbf{87.66/80.12} \\
     & AT & 87.35 & 77.74/66.37 & 79.15/67.34 & 76.87/65.51 & 81.23/68.13 \\
     & UniG-AT & \textbf{87.36} & \textbf{83.87/83.87} & \textbf{84.61/84.32} & \textbf{79.50/79.50} & \textbf{84.74/84.74} \\
    \midrule
    \multirow{4}{*}{ImageNet} & Vanilla & 78.47 & 59.79/14.91 & 52.71/6.70 & 50.30/7.14 & 41.59/7.30 \\
     & UniG & \textbf{78.47} & \textbf{64.35/51.79} & \textbf{66.14/52.88} & \textbf{55.95/38.21} & \textbf{56.97/40.41} \\
     & AT & \textbf{68.41} & 62.25/53.77 & 62.25/51.92 & 55.14/41.63 & 55.07/39.61\\
     & UniG-AT & 68.36 & \textbf{65.35/65.08} & \textbf{65.49/63.71} & \textbf{62.21/59.74} & \textbf{61.66/59.68} \\
  \bottomrule
\end{tabular}
\end{table}

\subsection{Adaptive Attack}

In general, an adaptive attack refers to an elaborate white-box attack with full knowledge of the defense strategy. Nevertheless, in the real-world cases, the attacker and the victim model are double-blind to each other, which means the defense strategy is not prior information for the attacker. Therefore, we consider another two adaptive attacks for black-box conditions: model stealing for transferable attacks \cite{kariyappa2021maze,orekondy2019knockoff} and tuning hyper-parameters for optimal attacks \cite{RND}.

The first one could utilize model outputs to estimate a surrogate model and then use the white-box adversarial examples of the surrogate model to attack the original model based on attack transferability \cite{papernot2017practical}. We here adopt two classical and practical steal-based attacks, eg. MAZE\footnote{https://github.com/sanjaykariyappa/MAZE} \cite{kariyappa2021maze} and KnockoffNet\footnote{https://github.com/tribhuvanesh/knockoffnets} \cite{orekondy2019knockoff}, to verify the effectiveness of our method. We use the recommended hyper-parameters in \cite{kariyappa2021maze}, and the query budget is set to $3\times10^7$.  Table \ref{model steal} demonstrates that UniG still outperforms baseline and RND by $>20\%$ under strong KnockoffNet attack. Although DENT outperforms UniG under the KnockoffNet attack, its performance under the more practical MAZE attack, which does not require a surrogate dataset as KnockoffNet does, is even worse than the baseline model, while our UniG improves the robust accuracy by $20.2\%$.

For the optimal hyper-parameter attack, we follow the settings in RND \cite{RND} and choose different hyper-parameters of Square, NES and Bandits attacks to find the optimal one. The result is shown in Table \ref{square}, \ref{nes&bandits}. As the square size or update step increases, our defense robustness decreases slightly, but within the normal fluctuation range as demonstrated by the performance of RND.

\vspace{5pt}
\begin{minipage}{\textwidth}
\begin{minipage}[t]{0.47\textwidth}
\centering
\makeatletter\def\@captype{table}
\caption{Remaining accuracy of RND, DENT, and UniG methods on CIFAR10 under steal-based adaptive attacks.}
\label{model steal}
\begin{tabular}{c|c|c|c}
    \toprule
     & Clean & MAZE & KnockoffNets \\
    \midrule
    Baseline & 94.26 & 65.00 & 21.95 \\
    RND & 90.91 & 84.05 & 30.02\\
    DENT & 94.25 & 59.06 & \textbf{64.14}\\
    UniG & \textbf{94.26} & \textbf{85.20} & 50.28 \\
    \bottomrule
\end{tabular}
\end{minipage}
\quad
\begin{minipage}[t]{0.47\textwidth}
\centering
\makeatletter\def\@captype{table}
\caption{Remaining accuracy of RND and UniG \emph{w.r.t.} square size of Square attack (query=100/2500) on CIFAR10.}
\label{square}
\begin{tabular}{c|c|c}

\toprule
square size & RND & UniG\\
\midrule
0.05 & 65.04/51.22 & \textbf{81.90/77.80}\\
0.1 & 63.25/47.64 & \textbf{80.99/74.89}\\
0.2 & 62.01/42.89 & \textbf{79.64/72.16}\\
0.3 & 60.83/41.05 & \textbf{78.79/70.11}\\
\bottomrule
\end{tabular}
\end{minipage}
\end{minipage}

\begin{table}[h!]
    \centering
    \caption{Remaining accuracy of RND / UniG \emph{w.r.t.} update step of NES / Bandits (query=100/2500) on CIFAR10.}
    \label{nes&bandits}
    \resizebox{\textwidth}{!}{
    \begin{tabular}{c|c|c|c|c|c|c|c}
        
        \toprule
        Attack & update step & 0.0001 & 0.0005 & 0.001 & 0.005 & 0.01 & 0.05 \\
        \midrule
        \multirow{2}{*}{NES} & RND & 87.49/79.29 & 86.58/78.38 & 86.58/78.38 & 85.79/\textbf{71.09} & 83.85/\textbf{68.36} & 74.73/59.24 \\
        & UniG & \textbf{94.26/91.43} & \textbf{94.26/86.72} & \textbf{93.32/82.95} & \textbf{90.49}/69.75 & \textbf{87.66}/66.92 & \textbf{79.18/65.98} \\
        \midrule
        \multirow{2}{*}{Bandits} & RND & 79.78/61.07 & 77.85/58.90 & 74.09/58.63 & 68.77/58.19 & 67.39/57.83 & 65.36/58.15 \\
        & UniG & \textbf{94.03/86.15} & \textbf{89.45/82.34} & \textbf{83.78/78.79} & \textbf{80.34/75.95} & \textbf{80.34/75.94} & \textbf{78.48/76.47} \\
        \bottomrule
    \end{tabular}}
\end{table}

\subsection{Hyper-parameter Study}
\begin{figure}
    \centering
    \subfigure[$\alpha$]{
    \includegraphics[width=0.22\textwidth]{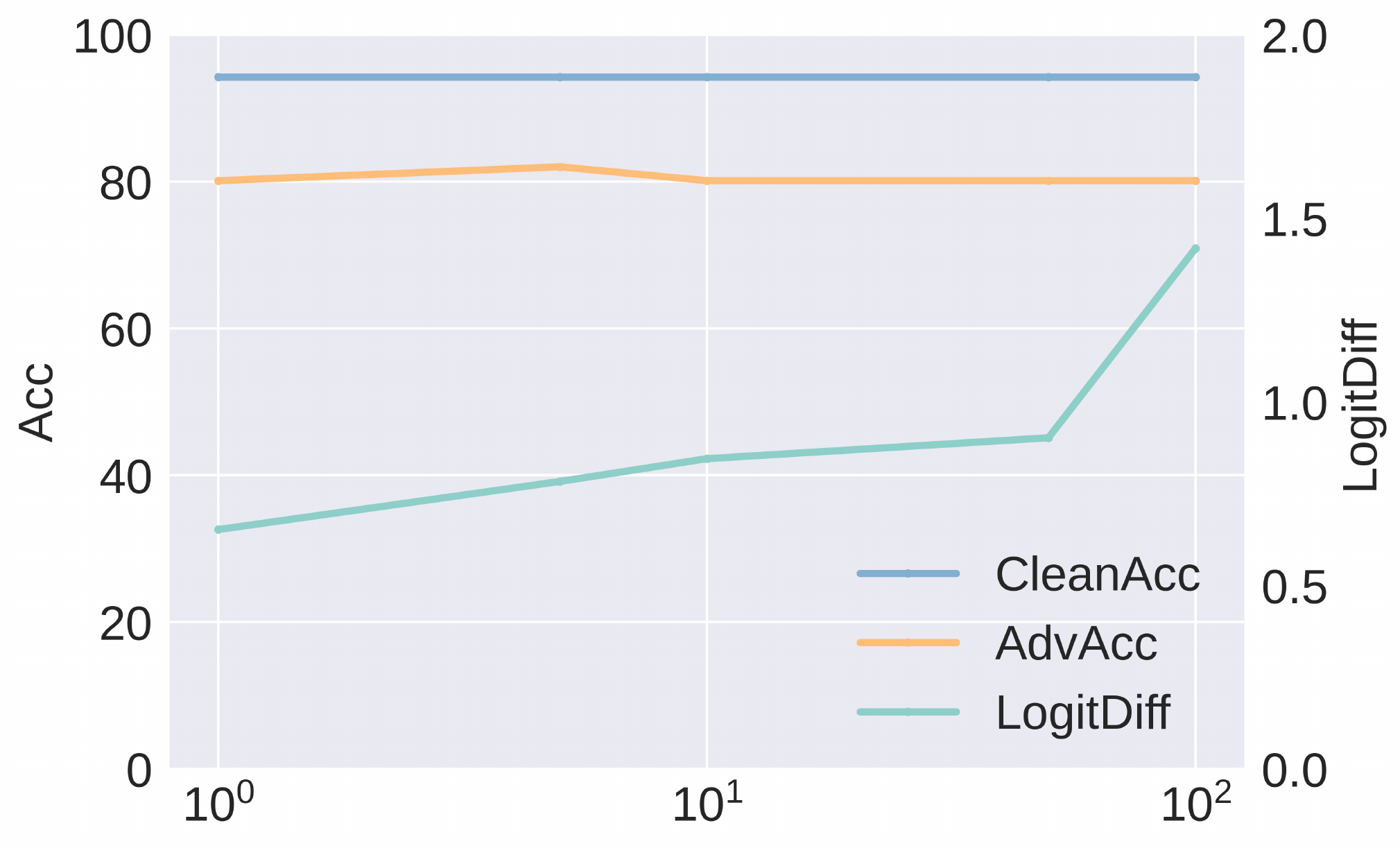}
    \label{lr-influence}
    }
    \subfigure[$p$]{
    \includegraphics[width=0.22\textwidth]{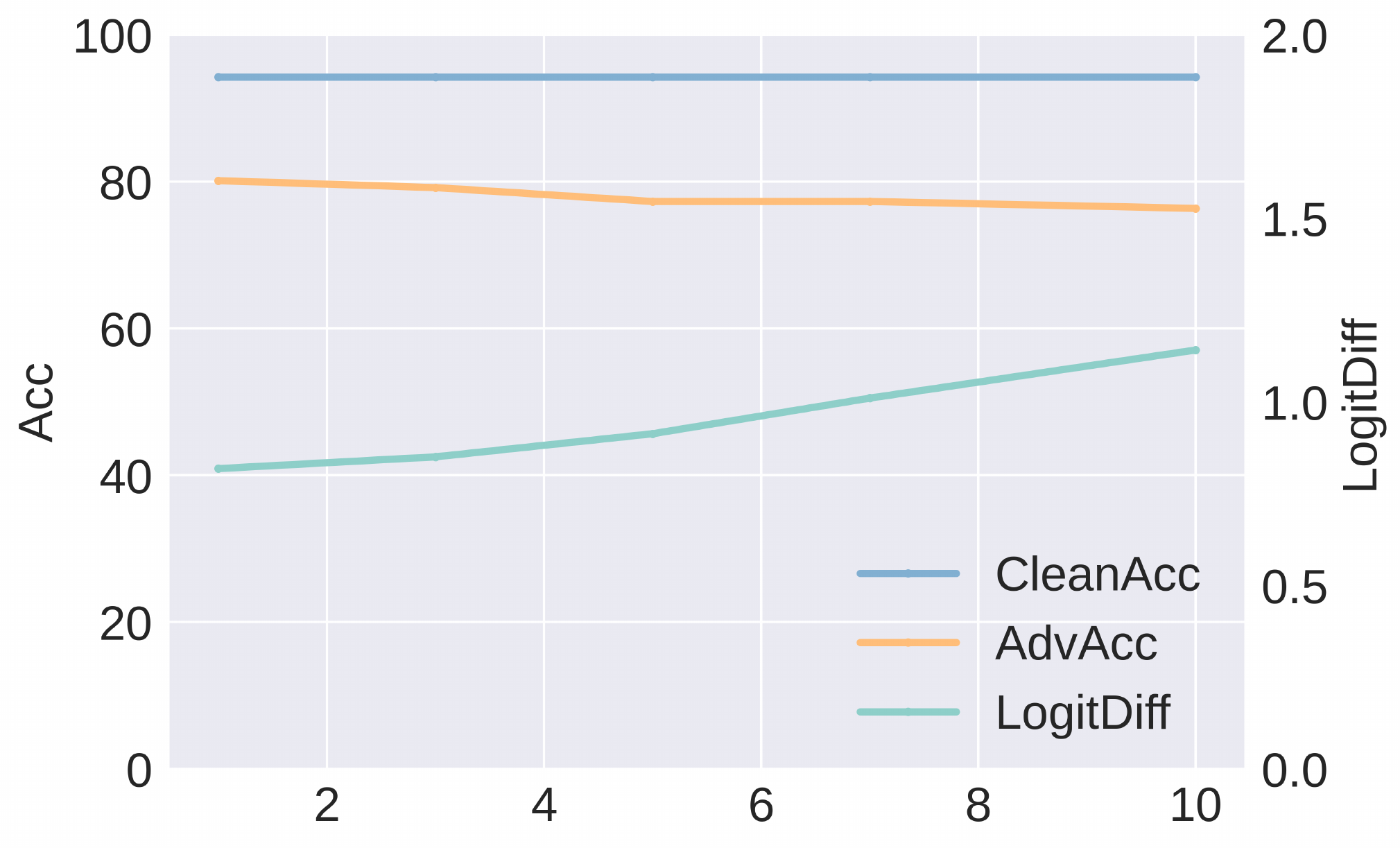}
    \label{epoch-influence}
    }
    \subfigure[$\delta$]{
    \includegraphics[width=0.22\textwidth]{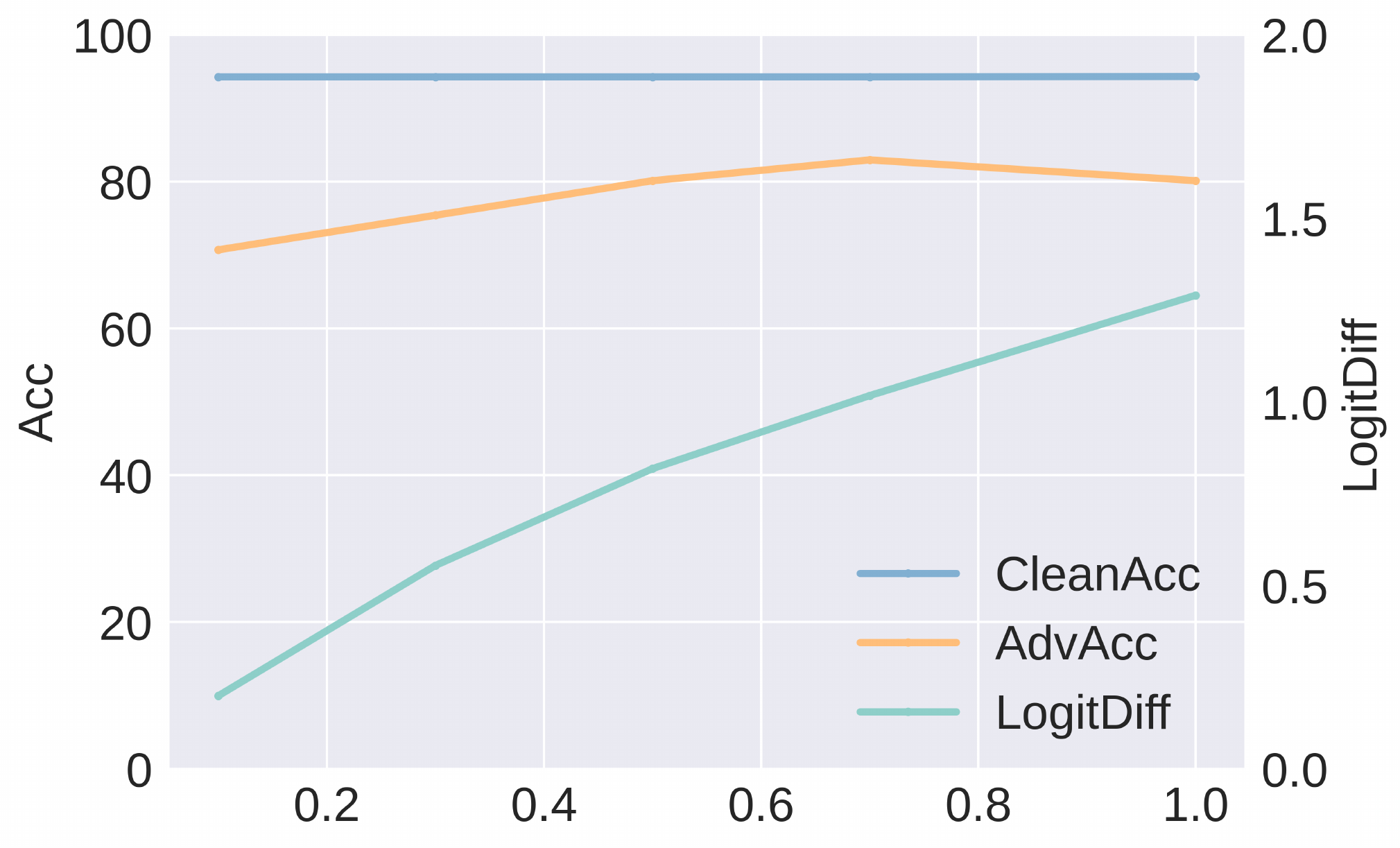}
    \label{delta-influence}
    }
    \subfigure[batch-size]{
    \includegraphics[width=0.22\textwidth]{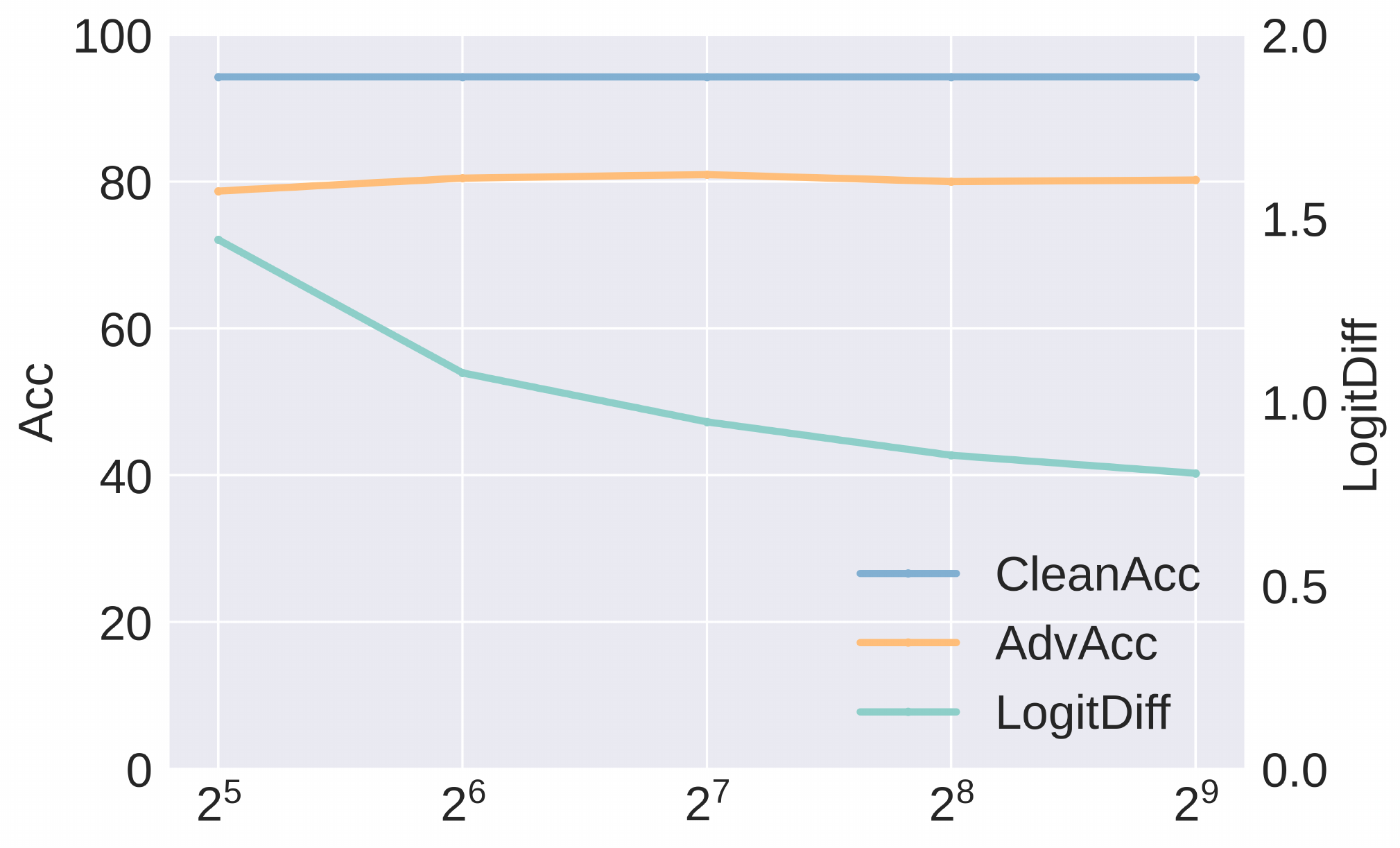}
    \label{batch-influence}
    }
    \caption{The influence of hyper-parameters on our method. The main hyper-parameters of UniG are the forward constraint parameter $\delta$, the optimization iterations $p$, and learning rate $\alpha$, and the batch size $b$, which influences our objective loss in batch-wise optimization.}
\end{figure}

Previous experiments are conducted using the fixed hyper-parameters as introduced before. To further evaluate our method, we here study the influence of different hyper-parameters on the performance of our method. The main hyper-parameters include the forward constraint parameter $\delta$, the optimization iteration $p$, and the learning rate $\alpha$. The parameter $\delta$ decides the largest element-wise difference between the parameters of our module and all-one matrix, which controls the trade-off between the clean performance and the robustness of our model. The iteration $p$ and the learning rate $\alpha$ are related to the optimization process of our module $A$. For fast test speed, we adopt one iteration and relatively bigger $\alpha$ to optimize $A$ in the previous experiments, while in this section, we adopt different $p$ and $\alpha$ to observe their influence on clean accuracy, logit difference and robustness under Square attack. As Fig \ref{lr-influence} shows, with the increase of $\alpha$, the logit difference slightly increases, but is still less than which of other methods.
And for the optimization iteration $p$, we observe the same phenomenon as $\alpha$, see Fig \ref{epoch-influence}. For the change of $\delta$, the robustness keeps stable, while the clean accuracy suffers from small fluctuations and the logit difference slightly increases, see Fig \ref{delta-influence}. Apart from the aforementioned hyper-parameters, we also evaluate our performance under different batch sizes, since our objective loss is dependent on the input test data. The result in Figure \ref{batch-influence} reveals that our method performs better with a larger batch size, but even with the batch size equals to 32, our robustness performance is quite good.


\subsection{Other Algorithmic Discussions\label{Other Algorithmic Discussions}}
This section discusses the computational complexity and the single test sample situation of our method. The first one is related to the time and space overhead of our method. And the single test sample condition is necessary to study taking the reality into account.

\textbf{Computational Complexity.} Because of the optimization process contained in our model's forward calculation, the computational complexity of our approach inevitably increases. There are two metrics to measure this complexity --- the number of model parameters and the number of Floating Point Operations (FLOPs) of the forward process. The first one reflects the space overhead and the second one represents the time overhead. Table \ref{time-consuming} reports the computational complexity of Vanilla, DENT and UniG model on CIFAR10 with the PreResNet18 architecture. Although there is some growth in the number of network parameters, the FLOPs of our method is far less than that of DENT, with a slight increase compared with that of the vanilla model.
\begin{table}[!t]
  \caption{The model complexity of vanilla, DENT and UniG network. High FLOPs indicate high complexity.}
  \label{time-consuming}
  \centering
  \begin{tabular}{c|c|c}
    \toprule
     Methods & Parameter (M) & FLOPs (GMac)\\
    \midrule
    Vanilla & 11.17 & 0.54 \\
    DENT & 11.17 & 7.78 \\
    UniG & 11.43 & 0.55 \\
  \bottomrule
  \end{tabular}
\end{table}

\textbf{Single Test Sample.} Since it is possible that some times the network receives a single image, it is imperative to discuss our performance in this case. Although our UniG, as well as most transduction defenses like DENT, are based on batch optimization, we propose a solution that cascading several (we use ten in our experiments) training data with test data to perform such optimization for the single test sample situation. Based on the experiment result on CIFAR10 in Table \ref{one test sample}, we can conclude that although there is a debasement of robustness when batch size equals to one, our method still surpasses baseline and DENT, and even slightly outperforms RND. We use Square attack and set $p=0.05$, query budget=100 here. Thus, although the primary application of our approach is not for the single test sample situation, our method can still improve robustness for this case.
\begin{table}[h!]
    \centering
    \caption{Remaining accuracy of DENT and UniG under Square attack (query=100) on CIFAR10 for the case of a single test sample. The last two columns are used to compare the performance.}
    \label{one test sample}
    \begin{tabular}{c|c|c|c|c|c}
        
        \toprule
        & Baseline & DENT(bz=1) & UniG(bz=1) & RND & UniG(bz=256)\\
        \midrule
        Clean & 94.26 & 83.22 & \textbf{94.26} & 91.14 & \textbf{94.26} \\
        Square & 38.79 & 5.32 & \textbf{72.24} & 65.04 & \textbf{81.90}\\
        \bottomrule
    \end{tabular}
\end{table}

\section{Conclusion}
In this paper, we propose a new defense method named Unifying Gradients (UniG) to defend against the most threatening attack in real applications---score-based query attacks (SQAs). The proposed method is based on the idea of distorting the gradient information contained in the query output by a slight modification on the forward output to fool the attacker into a weaker attack trajectory. In this paper, we choose the universal attack perturbation (UAP) as the weaker direction. Accordingly, the change on outputs is explicitly optimized with the gradient unification loss which indicates the UAP path. To practically implement this modification, we propose a Hadamard product operation module, which can be inserted into any pre-trained networks, and optimizes its parameter with the designed forward consistency and backward distortion loss at each inference time. With comprehensive experiments on CIFAR10 and ImageNet, it is verified that our approach can significantly boost the robustness under SQAs with no sacrifice of clean accuracy and a few variation on clean outputs. Noticing that the designed module is plug-and-play with negligible extra computational overhead, the overall method has a promising application prospect in the future.

\section*{Acknowledgement}
The authors are grateful to the anonymous reviewers for their insightful comments. This work was supported by National Natural Science Foundation of China (61977046), Shanghai Science and Technology Program (22511105600), and Shanghai Municipal Science and Technology Major Project (2021SHZDZX0102).

\bibliographystyle{ACM-Reference-Format}
\bibliography{main.bib}

\end{document}